\documentclass[11pt]{article}

\usepackage[preprint]{acl}

\usepackage{times}
\usepackage{latexsym}

\usepackage[T1]{fontenc}

\usepackage[utf8]{inputenc}

\usepackage{microtype}

\usepackage{inconsolata}

\usepackage{graphicx}

\usepackage{import}
\usepackage[inline]{enumitem}
\usepackage[singlelinecheck=off,font={footnotesize},labelsep=space]{caption}
\usepackage{float}
\usepackage{amsmath}
\usepackage{amssymb}
\usepackage[table,xcdraw]{xcolor}
\usepackage{subcaption}
\usepackage{flushend}
\usepackage{macros}

\title{Multi-LLM Collaboration for Medication Recommendation}

\author{
  Huascar Sanchez\textsuperscript{1}\thanks{*Equal contribution} \and
  Briland Hitaj\textsuperscript{1*} \and
  Jules Bergmann\textsuperscript{1,2} \and
  Linda Briesemeister\textsuperscript{1} \\
  \textsuperscript{1}Computer Science Laboratory, SRI International \\
  \texttt{\{huascar.sanchez, briland.hitaj, jules.bergmann, linda.briesemeister\}@sri.com} \\
  \textsuperscript{2}University of Maryland St. Joseph Medical Center
}

\begin{document}
\maketitle
\begin{abstract}
As healthcare increasingly turns to AI for scalable and trustworthy clinical decision support, ensuring reliability in model reasoning remains a critical challenge. Individual large language models (LLMs) are susceptible to hallucinations and inconsistency, whereas naive ensembles of models often fail to deliver stable and credible recommendations. Building on our previous work on \textit{LLM Chemistry}, which quantifies the collaborative compatibility among LLMs, we apply this framework to improve the reliability in medication recommendation from brief clinical vignettes.  Our approach leverages multi-LLM collaboration guided by Chemistry-inspired interaction modeling, enabling ensembles that are effective (exploiting complementary strengths), stable (producing consistent quality), and calibrated (minimizing interference and error amplification).
We evaluate our Chemistry-based Multi-LLM collaboration strategy on real-world clinical scenarios to investigate whether such interaction-aware ensembles can generate credible, patient-specific medication recommendations.
Preliminary results are encouraging, suggesting that LLM Chemistry-guided collaboration may offer a promising path toward reliable and trustworthy AI assistants in clinical practice.
\end{abstract}
\section{Introduction}
Medication recommendation from unstructured clinical notes remains a challenging task due to the high variability and ambiguity of patient narratives~\cite{eslami2025toward}. While large language models (LLMs) have demonstrated remarkable success across many clinical and biomedical applications, their performance is uneven---no single model consistently excels across reasoning, generation, and domain-specific understanding~\cite{chang2024survey}. This inconsistency makes medication recommendation error-prone, especially when relying on a single model’s inductive biases.

Recent efforts to improve reliability through model ensembling and routing~\cite{hu2024routerbench, liu2024dynamic, cloud2025subliminal} have focused primarily on selecting the best-performing individual models. However, these approaches often overlook the interaction dynamics between models --- how their reasoning processes reinforce or interfere with each other. As a result, ensembles can exhibit unreliable synergy, where collaboration amplifies not only strengths but also errors and biases.

To address this limitation, we introduce a Multi-LLM Collaboration approach that leverages the notion of LLM Chemistry~\cite{sanchez2025llmchemistry}---a quantitative measure of collaborative compatibility among LLMs. Our approach explicitly models the synergistic and antagonistic relationships that emerge when multiple models reason together, enabling the formation of ensembles that are: (1) Effective: leveraging complementary strengths to improve recommendation accuracy; (2) Stable: maintaining consistent performance across diverse clinical inputs; and (3) Calibrated:  minimizing interference and error amplification during collaboration, which translates to better task latency.

Building on this formulation, we evaluate how optimal multi-LLM collaboration can enhance reliability in clinical medication recommendation. Specifically, we investigate the following research question:~\emph{Does our approach ensure efficient, effective, stable, and calibrated multi-LLM collaboration for the medication recommendation use-case? If so, which LLM sampling strategy yields the most accurate results?}

While prior work has explored mixtures of LLMs --- by sequentially feeding one model's output into another to converge on a single aggregated result through majority voting~\cite{wang2024mixture}, by employing cascades to reduce costs where stronger models are invoked only when weaker ones fail~\cite{yue2024largelanguagemodelcascades}, by compensating for model deficiencies through meta-learning~\cite{zhou2024metagpt}, or by debating which model in a group is correct~\cite{chen2025optimizing}---we take a different approach. We introduce an inclusive and less restrictive collaboration framework in which any LLM, open-weight or proprietary, can participate. By leveraging diverse perspectives and complementary knowledge across models, our approach enhances answer consistency and mitigates individual model blind spots. This framework follows the intuition that LLMs should defend and refine their reasoning in group settings, where peer scrutiny fosters consensus on the quality and reliability of generated answers.

\noindent\textbf{Contributions.} In this paper, we build upon and extend our previous work~\cite{sanchez2025llmchemistry} to explore applications of multi-LLM collaboration in healthcare, leading to the following contributions:

\begin{itemize}[leftmargin=*,itemsep=2pt, parsep=0pt, topsep=0pt]
  \item Optimal Multi-LLM Collaboration. We leverage the concept of \textit{LLM Chemistry} to enable structured, interaction-aware, and optimal collaboration among multiple LLMs.
  \item Two-stage Collaboration Mechanism. We build on an existing multi-LLM collaboration setup and extend its evaluation step with a consensus step, transforming diverse, sometimes conflicting, model outputs into a unified, and reliable decision. 
  \item Application to Medication Recommendation. We evaluate the proposed approach on the clinical task of medication recommendation from patient notes, demonstrating improved accuracy and stability over the other ensemble baselines. 
\end{itemize}

\section{Background and Related Work}
\label{sec:background}

\textbf{Multi-LLM collaboration} has emerged as a promising strategy for improving robustness, reasoning diversity, and interpretability in LLM-based systems. Instead of relying on a single model, collaborative frameworks such as \textit{LLM ensembles}~\cite{du2023improving,yang2023one,chen2025harnessing}, \textit{Mixture-of-Agents (MoA)}~\cite{wang2024mixture,jang2025yalenlp}, \textit{Mixture-of-Domain Experts (MoDEM)}, and \textit{Mixture-of-Multimodal-Agents (MoMA)}~\cite{gao2025moma} combine multiple LLMs that exchange, critique, and refine outputs through structured interactions. Each model contributes distinct strengths---ranging from domain knowledge and factual precision to reasoning style and linguistic nuance---allowing the ensemble to achieve broader domain coverage and reliability than any individual component. Prior work has shown that multi-agent coordination enhances factuality and cross-domain reasoning through layered aggregation and verification mechanisms.

However, these collaborative strategies still exhibit key \textbf{limitations}~\cite{kassem2025robust,badawi2025can,li2025rethinking,chen2025harnessing,jain2025consensusmitigatingagreeablenessbias}. Most existing frameworks exhibit strong positive bias and rely on heuristic or task-specific aggregation without explicitly modeling the \textit{interaction dynamics} between participating models~\cite{hu2024routerbench}. As a result, collaboration outcomes often depend on implicit voting or averaging schemes that fail to distinguish when models reinforce versus contradict each other. This lack of interpretability makes it difficult to understand whether observed gains stem from genuine synergy or from redundancy among similar reasoning paths. Furthermore, multi-LLM ensembles tend to increase computational cost~\cite{ruiz2025wisdomdelusionllmensembles}, suffer from unstable scaling behavior across domains, and offer limited guidance on how to optimally pair or weight models during inference.

In the context of \textbf{healthcare}, where decisions require transparency, accuracy, and domain diversity, these weaknesses become critical. Tasks such as clinical summarization, diagnosis, multimodal data interpretation, or medicine recommendation demand fine-grained integration of heterogeneous expertise---from medical language understanding to quantitative reasoning over laboratory or imaging data. Existing multi-LLM systems like \textit{MoMA}~\cite{gao2025moma} address this partially by combining modality-specialized agents, yet they still rely on static ensembling mechanisms that overlook how model outputs interact or influence one another during joint reasoning. Without an explicit mechanism to assess these interactions, such systems risk inconsistent, unreliable outcomes across patient subgroups.

Our \textbf{Chemistry-based Multi-LLM approach}~\cite{sanchez2025llmchemistry} aims to overcome these limitations through a novel multi-LLM recommendation approach. This approach explicitly models the \textit{synergistic and antagonistic relationships} among collaborating LLMs. By quantifying and regulating these ``bonding'' effects, our framework captures how models complement or counteract each other, leading to more synergistic and reliable collaboration. This chemistry-inspired formulation provides a principled way to balance cooperation among LLMs, improving both computation efficiency, reasoning fidelity, and domain coverage---an especially vital property for healthcare applications where the cost of error is high and trustworthiness is paramount. This framework models collaboration as a two-stage process of response generation and evaluation, a common setup in
multi-LLM collaboration frameworks~\cite{du2023improving,madaan2023self,zhang2024chain}.

\subsection{Approach}
\label{sec:approach}

Our multi-LLM collaboration framework operates in two stages: \textbf{(1) Generation} and \textbf{(2) Evaluation}. These stages harness the complementary capabilities and independence of multiple LLMs to produce and verify medication recommendations from clinical notes. In the generation stage, a user request is distributed to a set of response generators, selected according to one of the sampling strategies described in Section~\ref{subsec:sampling}. The number of participating models $N$ is user-specified; in this work, we use $N=3$, the minimum required for majority agreement. Each LLM independently generates a medication recommendation, providing distinct reasoning paths that capture different linguistic, contextual, and domain biases. This variation forms the basis for cross-model evaluation and consensus formation in the next stage.

The evaluation stage performs quality control on the generated responses through a combination of \emph{review} and \emph{grading}. Each response is anonymously reviewed by other LLMs, ensuring balanced and independent assessment without contextual carry-over. Every review produces a \emph{grade} in $[0.0,1.0]$ reflecting the perceived accuracy, relevance, and completeness of the output. We reinterpret response generation as \textit{implicit evaluation}, since each generated output implicitly endorses its own validity. This unobtrusive mechanism reinforces quality assurance without additional overhead. By grading one another’s outputs, models collectively establish a measure of confidence in their recommendations. To compute aggregate quality and accuracy, we adopt a consensus-based estimation method inspired by the \textit{Vancouver crowdsourcing algorithm}~\cite{de2014crowdgrader} and detailed in our prior work~\cite{sanchez2025llmchemistry}. Together, these mechanisms identify high-quality, consensus-backed recommendations and quantify the reliability of each participating model, yielding consistent and trustworthy multi-LLM collaboration outcomes.

\section{Experimental Setup}
\label{sec:setup}

\subsection{Dataset Creation}
Our experiments use a dataset of brief clinical vignettes paired with corresponding medication recommendations. Each vignette is an unstructured note that concisely captures key information from a patient's medical record to facilitate communication among healthcare providers. In this paper, we use the terms \textit{clinical notes} and \textit{clinical vignettes} interchangeably. Each recommendation includes labeled fields describing the medication name, dosage, route of administration, frequency, timing (e.g., with meals), and the indication or condition treated. The fields for a single medication are linked to form a medication entry, and each recommendation contains multiple such entries. \emph{All medication entries were reviewed and validated by our team’s domain expert; entries that could not be confirmed were excluded from the dataset.}

The dataset comprises $20$ records synthesized by prompting LLMs to perform a synthesis task in reverse---generating plausible clinical vignettes for given sets of medications. Following the approach of Josifoski et al.~\cite{josifoski2023exploiting}, we exploit this asymmetry in task difficulty to produce high-quality pairs of clinical notes and medication recommendations. This strategy is particularly effective for our closed information task, where obtaining ground truth data is difficult, often inaccessible, or restricted by procedural barriers.

\subsection{LLMs Considered}

Table~\ref{tab:llms-used} lists the LLMs used in this study: ten proprietary models—OpenAI’s GPT~\cite{openaimodels}, Anthropic’s Claude~\cite{anthropicmodels}, and Google’s Gemini~\cite{googlemodels}—accessed through their APIs, and four open-source models accessed via Ollama~\cite{ollamamodels}. The selection spans diverse architectures, resource demands, and capabilities to enrich the resulting ensembles.

    \begin{table}
    \caption{Large Language Models (LLMs) Used }
    \centering
    \label{tab:llms-used}
    \resizebox{\columnwidth}{!}{%
    \begin{tabular}{|p{\columnwidth}|}
    \hline
    \rowcolor[HTML]{ECF4FF} 
    \textit{\textbf{Closed Source LLMs}} \\ \hline
claude-3-7-sonnet-20250219, 
claude-opus-4-1, 
claude-sonnet-4-5, 
gemini-2.0-flash, 
gemini-2.5-flash, 
gpt-4o, 
gpt-5, 
o1-mini, 
o3-mini, and
o4-mini,
    \\ \hline
    \rowcolor[HTML]{ECF4FF} 
    \textit{\textbf{Open Source LLMs}} \\ \hline
    firefunction-v2,
    gpt-oss:20b, 
    qwen3:32b, and
    qwen2.5:32b
    \\ \hline
    \end{tabular}
    }
    \end{table}

\subsection{LLM Sampling Strategies} %
\label{subsec:sampling}
In line with prior work considering the topic of multi-LLM collaboration, we consider the following LLM sampling strategies, i.e., mechanisms for defining the composition of an LLM ensemble:
\begin{enumerate*}[label=(\arabic*)]
    \item REMOTE - closed-source models (e.g., GPT, Claude, and Gemini model families),
    \item LOCAL - open-weights models,
    \item RANDOM - randomly selected models (sampled from both REMOTE and LOCAL sets), and
    \item CHEMISTRY (\textbf{ours}) - models recommended by our LLM Chemistry framework.
\end{enumerate*}

The first three strategies follow common practices in the literature and approximate realistic deployment scenarios for multi-LLM systems. In contrast, the \textbf{CHEMISTRY}-based strategy aims to identify the optimal subset of LLMs that exhibit strong collaborative compatibility (\emph{LLM Chemistry}) for the target task---\textbf{recommending medications from clinical notes for a specific patient}. Further details on the \textit{LLM Chemistry} formulation are provided in our prior work~\cite{sanchez2025llmchemistry}.

\subsection{Metrics}
We evaluate the proposed multi-LLM recommendation of medical prescriptions based on the following set of metrics:
\begin{enumerate*}[label=(\arabic*)]
    \item Efficiency: capturing the elapsed processing time for each LLM in a human readable format,
    \item Effectiveness: capturing the accuracy (i.e., effectiveness) of selection strategies in recommending medical prescriptions using the input (synthetic) clinical notes,
    \item Stability: capturing the ability of the selection strategies to maintain consistent performance across task, i.e., quality of results, and
    \item Calibration: capturing the degree of alignment among LLMs within an ensemble, i.e., the ability to avoid interference during collaboration.
\end{enumerate*}

\subsection{Protocol}

We evaluate 4 sampling strategies using LLM configurations (ensembles) composed of 3-models, i.e., $N = 3$. We execute each sampling strategy (LOCAL, REMOTE, and RANDOM) across 10-records of patient clinical notes, resulting in $\approx90$ answers, i.e., recommended prescriptions across all configurations. Note that the exact number varies due to likely execution failures. This is primarily evident in REMOTE and RANDOM sampling strategies, where communication with the remote models (e.g., GPT-*, or claude-*) may fail to complete. Using the results obtained from running LOCAL, REMOTE, and RANDOM sampling strategies, we then feed such results to our proposed LLM-Chemistry approach obtaining recommendations for LLM-ensembles of $N = 3$ exhibiting strong chemistry for the task of medical prescription recommendations. Under the CHEMISTRY selection strategy, we then conduct 10 independent trials per LLM ensemble size, adding $\approx300$ additional answers. Performance metrics are aggregated across tasks within each trial.

    \begin{figure}
		\includegraphics[width=0.9\linewidth]{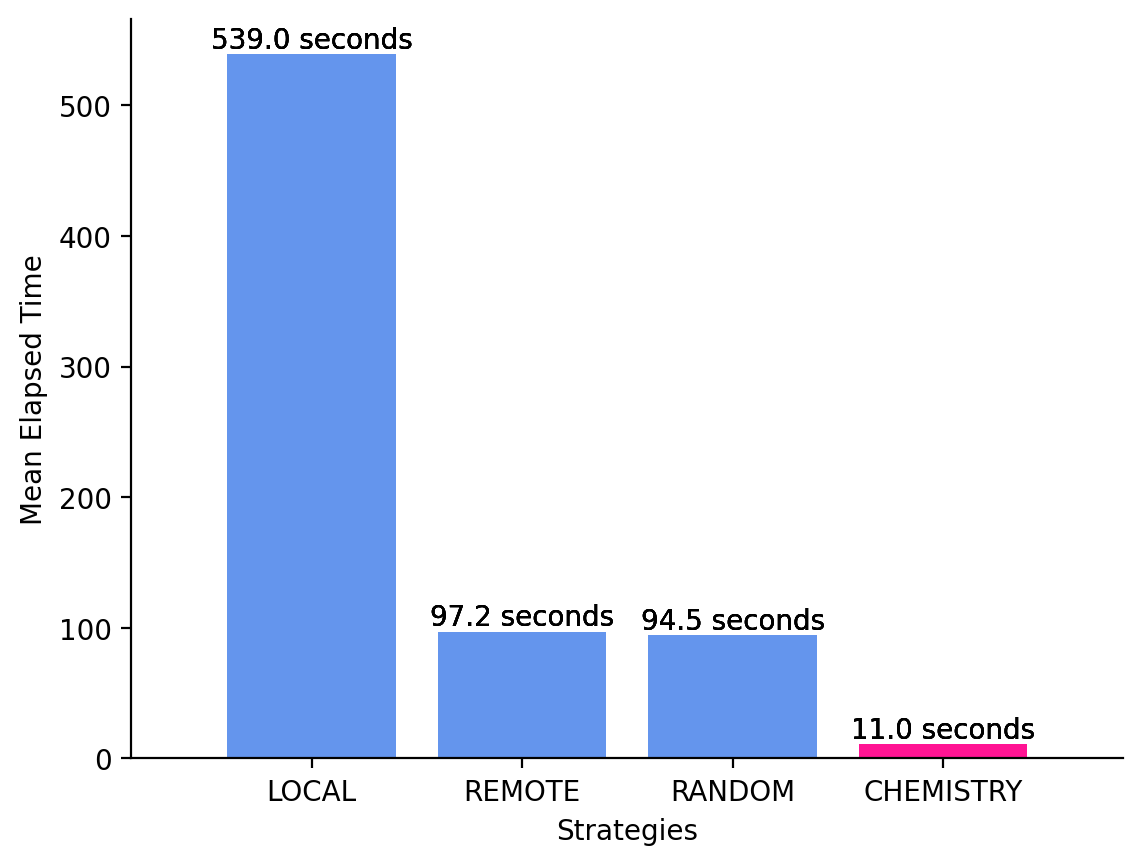}
		\caption{Efficiency comparison across sampling strategies. The CHEMISTRY-based multi-LLM ensemble, comprised of Claude models achieved an average generation time of 11 seconds, making it almost 9x faster than the nearest strategy (RANDOM), and nearly 49x faster than LOCAL-only ensembles when recommending medical prescriptions from brief clinical vignettes.}
		\label{fig:mean-elapsed-time}
	\end{figure}

\section{Experimental Results}
\label{sec:evaluation}

As part of our experimental evaluation synthetic clinical notes data, we evaluated our LLM Chemistry-based multi-LLM collaboration approach, within the medical prescription recommendation domain. The evaluation examined four dimensions–efficiency, effectiveness, stability, and calibration of multi-LLM (AI) collaboration–to assess the performance of the Chemistry-based multi-LLM strategy,~\ref{sec:approach}. In this section, we present (and describe) the results for each metric.

    \subsection{Efficiency}

During our experiments, we recorded the elapsed processing time for each LLM execution in a human readable format. For example, the ``o3-mini'' model, executed using the REMOTE sampling strategy, required 3 minutes and 37 seconds to generate a medical prescription recommendation for a given clinical note. As a reminder, the CHEMISTRY-based strategy (ours) aims to select the most suitable ensemble of LLMs for a given task, striving not only for high-quality outputs but also for efficiency in resource utilization–considering both the models employed and the time required for generation. Such efficiency is essential for the practical adoption of multi-LLM approaches in medical contexts.

Figure~\ref{fig:mean-elapsed-time} illustrates the efficiency results. Using the information generated by the LOCAL, REMOTE, and RANDOM sampling strategies for the current task, the LLM CHEMISTRY framework identified a three-model ensemble as the most effective configuration. The selected models were: 1) claude-3-7-sonnet-20250219, 2) claude-opus-4-1, and 3) claude-sonnet-4-5. This ensemble, referred to as the \textbf{CHEMISTRY} ensemble in this paper, was evaluated across 10-clinical notes, each executed 10 times, and the mean elapsed generation time was computed. On average, the CHEMISTRY ensemble produced recommendations in 11 seconds, which is approximately nine times faster than the RANDOM (94.5s) and REMOTE (97.2s) strategies, and nearly 49-times faster than ensembles composed exclusively of LOCAL LLMs. This demonstrates a clear efficiency advantage of the CHEMISTRY-based multi-LLM approach in medical recommendation use cases.

    \subsection{Effectiveness}
We next evaluate the effectiveness (i.e., the accuracy) of CHEMISTRY-based multi-LLM strategy in recommending medical prescriptions for a set of clinical notes.

    \begin{figure}
		\includegraphics[width=0.9\linewidth]{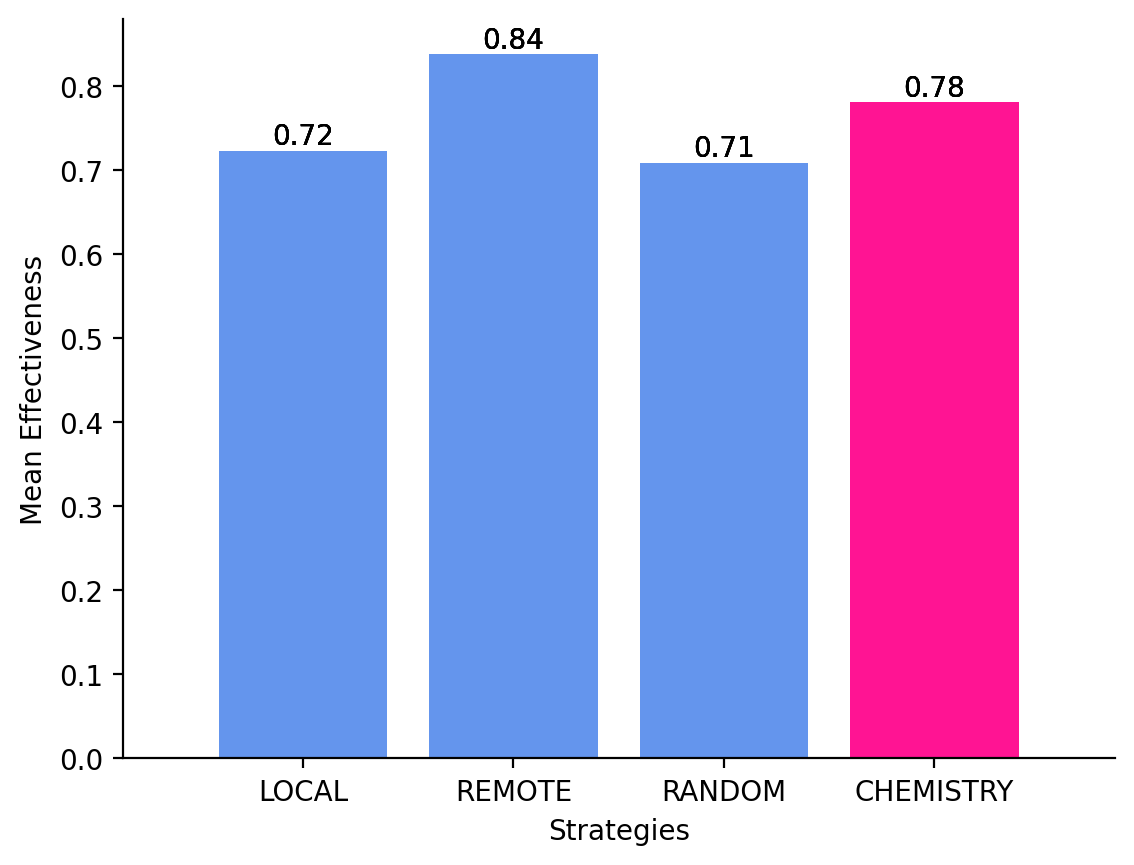}
		\caption{Effectiveness comparison across sampling strategies. The CHEMISTRY ensemble achieved an accuracy of 0.78, closely matching the REMOTE strategy (0.84) while outperforming other strategies. The ensemble, composed of Claude models from Anthropic, balances both accuracy and efficiency in medical prescription recommendations.}
		\label{fig:mean-effectiveness-per-strategy}
	\end{figure}

In our experiments, the CHEMISTRY ensemble achieved an accuracy of 0.78, comparable to the REMOTE ensembles (0.84), while outperforming all other sampling strategies,~Figure~\ref{fig:mean-effectiveness-per-strategy}. Notably, the CHEMISTRY ensemble consists of only remote models–specifically, Anthropic's Claude models–and incorporates additional factors such as efficiency, as discussed earlier in this section. These results indicate that CHEMISTRY-based multi-LLM ensemble maintains competitive accuracy while optimizing for other key metrics such as efficiency.

    \subsection{Stability}

Stability refers to CHEMISTRY-based ensembles ability to maintain consistent performance across multiple tasks–in this case, its capacity to generate high-quality medical prescription recommendations reliably. 
    \begin{figure}
		\includegraphics[width=0.9\linewidth]{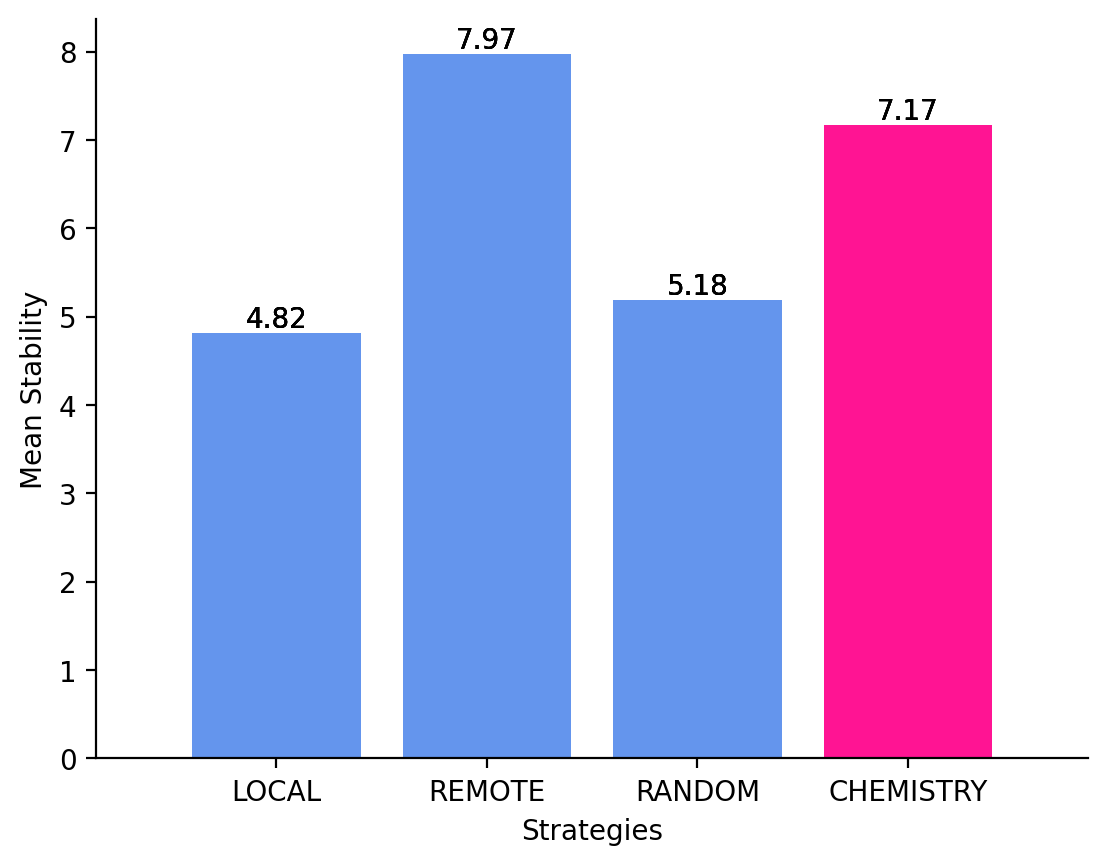}
		\caption{Stability comparison across sampling strategies. The CHEMISTRY ensemble maintained stability comparable to REMOTE ensembles while surpassing LOCAL and RANDOM strategies. Unlike the REMOTE strategy, which exhibited occasional execution failures, the CHEMISTRY strategy showed no failures, providing evidence of its robustness and reliability.}
		\label{fig:mean-stability-per-strategy}
	\end{figure}
In our experiments, the CHEMISTRY ensemble demonstrated stability comparable to multi-LLM ensembles composed of REMOTE models, while significantly outperforming both LOCAL and RANDOM counterparts, Figure~\ref{fig:mean-stability-per-strategy}. 
While failures were observed in some REMOTE strategy executions, no such failures occurred in the CHEMISTRY ensemble, further underscoring the stability and robustness of its performance. These findings highlight CHEMISTRY-based multi-LLM ensemble's ability to deliver consistent and dependable results across repeated medical recommendation tasks.

    \subsection{Calibration}

Calibration refers to the degree of alignment among LLMs within a multi-LLM ensemble --- that is, the ability to avoid interference and error amplification during collaboration. Well-calibrated ensembles maintain consistent confidence and decision criteria, ensuring that differences among models reflect genuine informational diversity rather than noise or overconfidence.

\begin{figure}
		\includegraphics[width=0.9\linewidth]{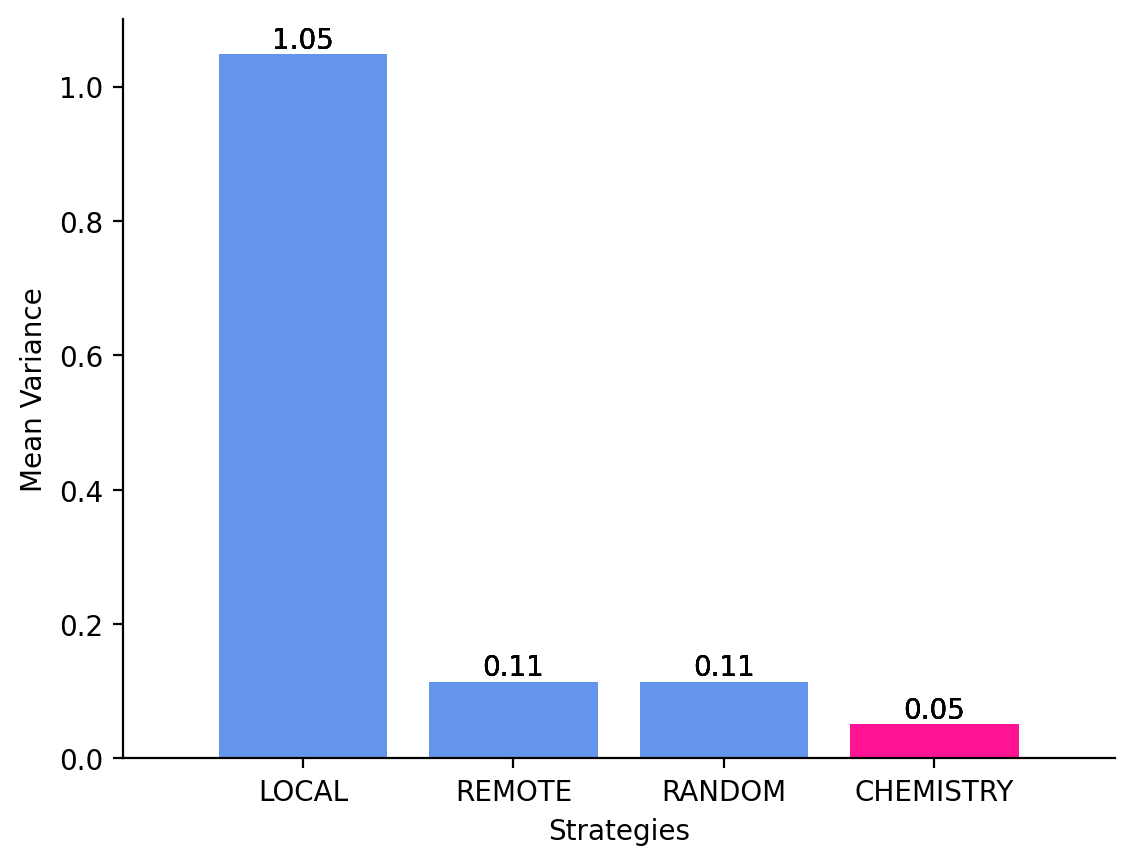}
		\caption{Calibration comparison across sampling strategies. The CHEMISTRY ensemble achieved the lowest variance ($0.05$), indicating high inter-model agreement and effective calibration. By contrast, REMOTE and RANDOM ensembles exhibited moderate variance ($0.11$), while LOCAL ensembles showed poor calibration ($1.05$), reflecting weak consensus among constituent LLMs.}
		\label{fig:mean-variance-per-strategy}
	\end{figure}

To evaluate this property, we measured the mean variance of agreement across sampling strategies (Figure~\ref{fig:mean-variance-per-strategy}). The CHEMISTRY ensemble achieved a mean variance of 0.05, substantially lower than REMOTE (0.11) ensembles, and far below LOCAL ensembles (1.05). This indicates that models selected via the CHEMISTRY strategy are highly calibrated and maintain strong internal consensus when generating medical prescription recommendations.
    
These findings suggest that calibration --- serving as a precondition for agreement --- enables the CHEMISTRY strategy to reduce noise and error propagation, ensuring that consensus among LLMs emerges from reliable, evidence-aligned reasoning.

\subsection{Sample Outputs}

In~Figure~\ref{fig:sample-outputs}, we provide snippets of recommendations generated by the multi-LLM ensembles formed under our \textbf{CHEMISTRY} strategy. 
All experiments were conducted in a zero-shot setting using synthetic but domain expert–validated clinical notes and recommendations to ensure methodological rigor and reproducibility.

\begin{figure}[ht]
    \centering
    \small
    \hspace{-2em}
    \begin{subfigure}[b]{.94\columnwidth}
        \centering
        \begin{tabular}{|p{\textwidth}|}
        \hline
        \rowcolor[HTML]{EFEFEF} 
        \textbf{Clinical Vignette:}~Jane Doe is a 72yo female who has diabetes, high cholesterol, heart disease, and chronic pain in her knees. She has trouble taking her medications and no other concerns. \\ \hline
        \textbf{Medication Recommendation:} 
        \begin{itemize}[leftmargin=*,itemsep=0pt, parsep=0pt, topsep=0pt]
            \item Metformin 500 mg PO BID with meals (diabetes).
            \item Atorvastatin 20 mg PO daily in evening (high cholesterol).
            \item Lisinopril 10 mg PO daily (heart disease, diabetes with renal protection).
            \item Acetaminophen 650 mg PO q6h PRN (chronic knee pain).
            \item Furosemide 20 mg PO daily in morning (heart disease).
        \end{itemize}
        \\ \hline
        \end{tabular}
    \end{subfigure}
    \vspace{0em}
    \hspace{-2em}
    \begin{subfigure}[b]{.94\columnwidth}
        \centering
        \begin{tabular}{|p{\textwidth}|}
        \hline
        \rowcolor[HTML]{EFEFEF} 
        \textbf{Clinical Vignette:}~Patient is a 76-year-old white female with a medical history of hypertension, osteoarthritis, osteoporosis, hypothyroidism, allergic rhinitis, and kidney stones. She presents today for a refill of her blood pressure medication. She reports no new concerns, changes in her chronic conditions, or symptoms such as fever, chills, cough, congestion, nausea, vomiting, chest pain, or chest pressure. \\ \hline
        \textbf{Medication Recommendation:} 
        \begin{itemize}[leftmargin=*,itemsep=0pt, parsep=0pt, topsep=0pt]
            \item Lisinopril 10 mg daily PO (hypertension).
            \item Hydrochlorothiazide 25 mg daily PO (hypertension).
            \item Amlodipine 5 mg daily PO (hypertension).
            \item Levothyroxine 50 mcg daily PO (hypothyroidism).
            \item Cetirizine 10 mg daily PO (allergic rhinitis).
        \end{itemize}
        \\ \hline
        \end{tabular}
    \end{subfigure}
    \vspace{0.5em}
    \hspace{-2em}
    \begin{subfigure}[b]{.94\columnwidth}
        \centering
        \begin{tabular}{|p{\textwidth}|}
        \hline
        \rowcolor[HTML]{EFEFEF} 
        \textbf{Clinical Vignette:}~The 76-year-old white female patient presented for a blood pressure medication refill. She has a documented history of hypertension, osteoarthritis, osteoporosis, hypothyroidism, allergic rhinitis, and kidney stones, as noted during her previous consultation with Doctor Kumar. No new symptoms or changes in her condition were reported at this visit. \\ \hline
        \textbf{Medication Recommendation:} 
        \begin{itemize}[leftmargin=*,itemsep=0pt, parsep=0pt, topsep=0pt]
            \item Lisinopril 10 mg PO daily (hypertension).
            \item Levothyroxine 50 mcg PO daily (hypothyroidism).
            \item Alendronate 70 mg PO weekly (osteoporosis).
            \item Acetaminophen 500 mg PO every 6 hours as needed (osteoarthritis).
            \item Cetirizine 10 mg PO daily (allergic rhinitis).
        \end{itemize}
        \\ \hline
        \end{tabular}
    \end{subfigure}    
    \caption{Sample outputs produced by our CHEMISTRY-based Multi-LLM Recommendation approach addressing the following task: ``Recommend Necessary Medical Prescriptions''}
    \label{fig:sample-outputs}
\end{figure}
\section{Conclusions and Future Work}
\label{sec:conclusions}

This work demonstrated the feasibility of using \textbf{Chemistry}-based multi-LLM collaboration to generate and recommend medical prescriptions from brief clinical notes. Our results show that our approach enables efficient, effective, stable, and calibrated multi-AI collaboration. In our experiments, CHEMISTRY-recommended ensembles yielded high-quality prescription recommendations and consistently matched or outperformed other LLM sampling strategies (\textit{LOCAL}, \textit{REMOTE}, and \textit{RANDOM}) across these reliability dimensions.

\noindent\textbf{Future Work.} 
Our experiments were conducted on synthetic, domain expert–validated data containing limited patient information (e.g., brief clinical notes without accompanying clinical practice guidelines) and used a zero-shot prompting setup. Future work will extend evaluation to larger, real-world datasets that include richer patient context---such as detailed notes, current medications, dosages, and allergy information---to assess generalizability and clinical safety. Incorporating retrieval-augmented generation (RAG) capabilities into the Chemistry-based multi-LLM Collaboration approach will further allow ensembles to access up-to-date medical knowledge from clinical practice guidelines and recent publications, improving grounding and transparency. Despite these limitations, our findings indicate that multi-LLM collaboration guided by LLM Chemistry is a practical and reliable approach for healthcare-related tasks.
\section*{Acknowledgments}
This research is based upon work supported in part by the Advanced Research Projects Agency for Health (ARPA-H), Defense Logistics Agency (DLA) under Contract Number SP4701-23-C-0073. Any opinions, findings and conclusions or recommendations expressed in this material are those of the authors and do not necessarily reflect the views of Advanced Research Projects Agency for Health (ARPA-H), Defense Logistics Agency (DLA), or the United States Government. The U.S. Government is authorized to reproduce and distribute reprints for governmental purposes notwithstanding any copyright annotation therein.

\bibliography{main}

\end{document}